%% file: main.tex
\documentclass[letterpaper, 10 pt, conference]{ieeeconf}  

\IEEEoverridecommandlockouts                              

\usepackage{graphics} 
\usepackage{epsfig} 
\usepackage{mathptmx} 
\usepackage{times} 
\usepackage{amsmath} 
\usepackage{amssymb}  

\usepackage{graphicx}
\usepackage{subcaption}

\usepackage{booktabs}

\usepackage{float}
\usepackage{hyperref}

\title{\LARGE \bf
Meta-Learning Regrasping Strategies for Physical-Agnostic Objects
}

\author{Ning Gao$^{1, 2,\dagger}$ Jingyu Zhang$^{1}$ Ruijie Chen$^{1}$ Ngo Anh Vien$^{1}$ Hanna Ziesche$^{1}$ Gerhard Neumann$^{2}$
\thanks{\textdagger\ Corresponding author. Email: \tt\small ning.gao@de.bosch.com}
\thanks{$^{1}$ Bosch Center for Artificial Intelligence, Renningen, Germany.}%
\thanks{$^{2}$ Autonomous Learning Robots Lab, Karlsruhe Institute of Technology, Karlsruhe, Germany}%
}

\begin{document}

\maketitle
\thispagestyle{empty}
\pagestyle{empty}

\input{docs/abstract}
\input{docs/introduction}

\input{docs/related_work}
\input{docs/methodology}
\input{docs/experiments}
\input{docs/conclusion}


\clearpage
\bibliographystyle{IEEEtran}
\bibliography{IEEEabrv}




\end{document}

%% file: docs/abstract.tex
\begin{abstract}

Grasping inhomogeneous objects in real-world applications remains a challenging task due to the unknown physical properties such as mass distribution and coefficient of friction. In this study, we propose a meta-learning algorithm called ConDex, which incorporates Conditional Neural Processes (CNP) with DexNet-2.0 to autonomously discern the underlying physical properties of objects using depth images. ConDex efficiently acquires physical embeddings from limited trials, enabling precise grasping point estimation. Furthermore, ConDex is capable of updating the predicted grasping quality iteratively from new trials in an online fashion. To the best of our knowledge, we are the first who generate two object datasets focusing on inhomogeneous physical properties with varying mass distributions and friction coefficients. 
Extensive evaluations in simulation demonstrate ConDex's superior performance over DexNet-2.0 and existing meta-learning-based grasping pipelines. Furthermore, ConDex shows robust generalization to previously unseen real-world objects despite training solely in the simulation. The synthetic and real-world datasets will be published as well.
\end{abstract}

%% file: docs/introduction.tex
\section{INTRODUCTION}

Grasp detection is one of the fundamental problems in robotic manipulation, which has led to great progress in recent years thanks to the advancements of deep learning techniques.
Grasp detection is normally formed as finding the stable grasp position w.r.t. the geometry of the object, the configuration of the end-effector, and the specific manipulation tasks~\cite{review}. 
Extensive studies have investigated various end-effector such as parallel jaw~\cite{Mahler2016,Mahler2017}, suction gripper~\cite{suctionAmazon,Mahler2018} and multi-finger gripper~\cite{multifinger} using RGB-D~\cite{Jiang2011,Lenz2015} or depth~\cite{Morrison2020} images from synthetic~\cite{Schaub20206DOFGD} or real-scene~\cite{Song2020,graspnet-dataset, Levine2018} datasets, with the purpose of increasing the generalization on unseen objects~\cite{Kal2018}, closing the sim-to-real gap~\cite{Klee2020,Quillen2018}, and increasing the robustness against occlusion~\cite{vgn}. Recently, ~\cite{Huang2022DefGraspSimPS,Farias2022GraspTF,Rho2021LearningFF} further improve the performance on grasping deformable objects.

\begin{figure}[t]
	\centering
		\includegraphics[width=0.5\textwidth]{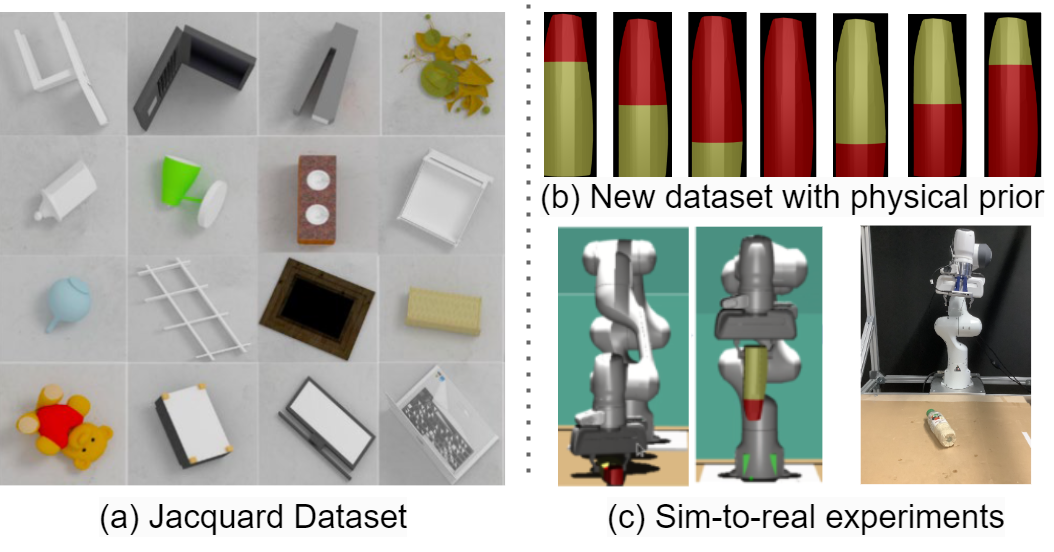}
	\caption{(a) Existing datasets such as Jacquard dataset ~\cite{Jacquard} exhibit a variety of textures and geometries. In contrast, (b) our research centers on heterogeneous physical properties (mass distribution and friction coefficients) across the object. For instance, the red part denotes higher mass density and yellow denotes lower mass density. (c) We further evaluate our method in a sim-to-real scenario.}
	\label{fig:intro}
\end{figure}

Although these methods have achieved promising results on various open datasets, it is noteworthy that these datasets predominantly adhere to a homogeneous assumption of the physical properties. 
For example, as shown in Fig.~\ref{fig:intro}a, the 3D objects~\cite{shapenet,Kasper2012, Jacquard} in simulation and 3D printed objects~\cite{Morrison2020_dataset} are typically treated as entire entities, neglecting the consideration of diverse material properties and friction coefficients for individual components, while most real-world datasets~\cite{Calli2015,BigBIRD,Lenz2013DeepLF,Young2009} frequently exhibit a variety of textures and geometries but tend to feature uniform distribution of mass. Crucially, neither the training nor evaluation stages explicitly incorporate physical properties. Thus, most vision-based grasp detection algorithms rely solely on geometries and textures.  This limitation becomes evident in practical scenarios, exemplified by an instance where methods relying solely on object geometry and texture fail to effectively lift an object due to the oversight of variations in part density or friction coefficients.

In contrast to prior works, we explicitly construct various objects with distinct mass distributions and friction coefficients as shown in Fig.~\ref{fig:intro}b and employ a shared model to acquire knowledge of these properties purely from depth images, emphasizing the significance of discerning physical attributes purely from visual input. 
We frame this challenge as a few-shot learning problem, i.e., wherein the physical properties of each object must be gleaned from contextual information derived from a limited number of grasp trials. 
In essence, our method aims to emulate human learning principles by: 
1) Accumulatively acquiring knowledge of physical properties through previous experiences.
2) Facilitating learning during both online and offline inference processes. 
3) Seamlessly integrating into existing grasp pipelines without compromising performance.
4) Enhancing real-world performance while leveraging knowledge gained from simulation.

Conditional Neural Processes (CNP)~\cite{Garnelo2018} has shown advances in few-shot classification and regression tasks~\cite{Gao_2022_CVPR}, characterized by rapid adaptation and inference capabilities. In our work, we integrate CNP into DexNet-2.0~\cite{Mahler2017} with minimal alterations to the original grasp frameworks and encode the object's physical properties as latent representations derived from contextual information. 
To address the limited availability of appropriate datasets, we create two synthetic datasets characterized by distinguishable physical properties in comparison to existing ones. Subsequently, we conduct performance evaluations using the Pybullet and Mujoco simulators, including novel objects from both intra-category (IC) and cross-category (CC). 
Furthermore, we extend our evaluation to real-world scenarios, where the model is exclusively trained in Mujoco, facilitating the investigation of the sim-to-real gap (shown in Fig.~\ref{fig:intro}c).

In summary, our contributions can be summarized as follows: 
\begin{itemize}
    \item We introduce a novel meta-learning grasp framework aimed at addressing the relatively unexplored challenge of grasping objects characterized by diverse physical properties, relying solely on visual input.
    \item We introduce two innovative synthetic datasets that explicitly incorporate physical properties, making them compatible with a wide range of simulation frameworks.
    \item Our approach demonstrates substantial advantages in real-world object manipulation, despite being trained exclusively in a simulated environment.
\end{itemize}

%% file: docs/related_work.tex
\section{RELATED WORK}





\subsection{Few-Shot Grasp Detection}
Few-shot learning is crucial to generic grasp detection, e.g., generalizing to unseen objects or layouts with grasp preference from only a few examples~\cite{Du2020VisionbasedRG}. DemoGrasp~\cite{DemoGrasp} reconstructs the object mesh and predicts the grasp pose from a sequence of RGB-D images with a human demonstration while ~\cite{Hlnon2020LearningPA} learns the grasp point from demonstrations including both authorised and prohibited locations. FSG-Net~\cite{FSG-Net} and GAS~\cite{Kaynar2023RemoteTG} employ a few-shot semantic segmentation module to grasp a specific object from the clutter although the object has never shown during training. Meanwhile, IGML~\cite{IGML}, LGPS~\cite{LGPS} and TACK~\cite{TACK} predict the grasp point of a novel object out of clutter given a few examples with the specified grasp point. 
Our work shares similarities with IGML by employing a meta-learning algorithm, however, our approach dispenses with the need for predefined grasp point labels and facilitates online adaptation with rapid inference capabilities.

\subsection{Grasp Datasets}
A variety of grasp datasets have been proposed over recent years. ~\cite{Berscheid2021RobotLO,softdata,MVGrasp} collect dataset using real robots while ~\cite{graspnet-dataset,Kal2018,Mahler2017} combine the real-world and simulated data. For example, GraspNet-1Billion~\cite{graspnet-dataset} captures real RGB-D images with generated grasp poses. Meanwhile, ~\cite{ACRONYM,DexGraspNet,DVGG} include purely the synthetic datasets based on simulation. 
Nonetheless, none of the previously mentioned datasets has provided explicit definitions of discernible physical attributes across distinct parts of individual objects. Conversely, the prevailing norm entails objects exhibiting uniform mass distribution and friction coefficients, whether within a simulated environment or in real-world scenarios.
For instance, DVGG~\cite{DVGG} shares identical friction coefficient (0.25) and density (1500 $\mathrm{kg/m^{3}}$) over all objects within the simulation. Similarly, widely-used large-scale object datasets like ShapeNet~\cite{shapenet} and ObjectNet3D~\cite{ObjectNet3D} lack explicit physical configuration, whereas commonly employed object sets like YCB~\cite{Calli2015}, KIT~\cite{Kasper2012} and BigBIRD~\cite{BigBIRD} predominantly feature household items or toys characterized by uniform mass distribution.

\subsection{Meta-Learning}
In the realm of meta-learning, a learning agent acquires meta-knowledge from previous learning episodes or different domains and then uses this acquired knowledge to improve the learning on future tasks~\cite{metalearning-review}. MAML~\cite{maml} is an optimization-based meta-learning method where meta-knowledge is encapsulated within model parameters, and adaptation to new tasks is achieved through further optimization during inference. 
In contrast, Neural Processes (NPs) fall within the category of neural latent variable models and interpret meta-learning as conditional few-shot function regression~\cite{NP}. Similar to Gaussian Processes, NPs model function distributions conditioned on contextual information~\cite{NP,Kim2019,VERSA}. 
Meta-learning algorithms have been applied in various domains, including low-dimensional function regression~\cite{Garnelo2018,NP,DSVNP}, image completion~\cite{CCNP,FNP,RNP}, and few-shot classification~\cite{relation-network, MMAML}.
Recent advancements~\cite{TaskAugmentation, Gao_2022_CVPR} have extended the application of meta-learning to pose estimation. Additionally, ~\cite{FRCL,Gao_2022_CVPR} enhance meta-learning by incorporating contrastive representation learning from disjoint context sets. CLNP~\cite{CLNP} further extends this idea to time series data by combining contrastive learning with ConvNP~\cite{CCNP}. In our work, we employ CNP~\cite{Garnelo2018} to meta-learn a latent embedding to represent the physical properties of each object which facilitates online adaptation and expedites the inference process. 

%% file: docs/methodology.tex
\section{META-GRASPING PHYSICAL-AGNOSTIC OBJECTS}


In this section, we present the creation of object assets characterized by diverse physical attributes, encompassing both \textit{Letters} datasets and \textit{Bottles} datasets. We introduce the process of gathering data from simulators utilizing Pybullet and Mujoco, as well as elucidate the meta-learner pipeline namely \textit{ConDex} to discern these physical properties.

\subsection{Datasets with Heterogeneous Physical Properties} 

\begin{figure}[t]
	\centering
		\centering
		\includegraphics[width=0.48\textwidth]{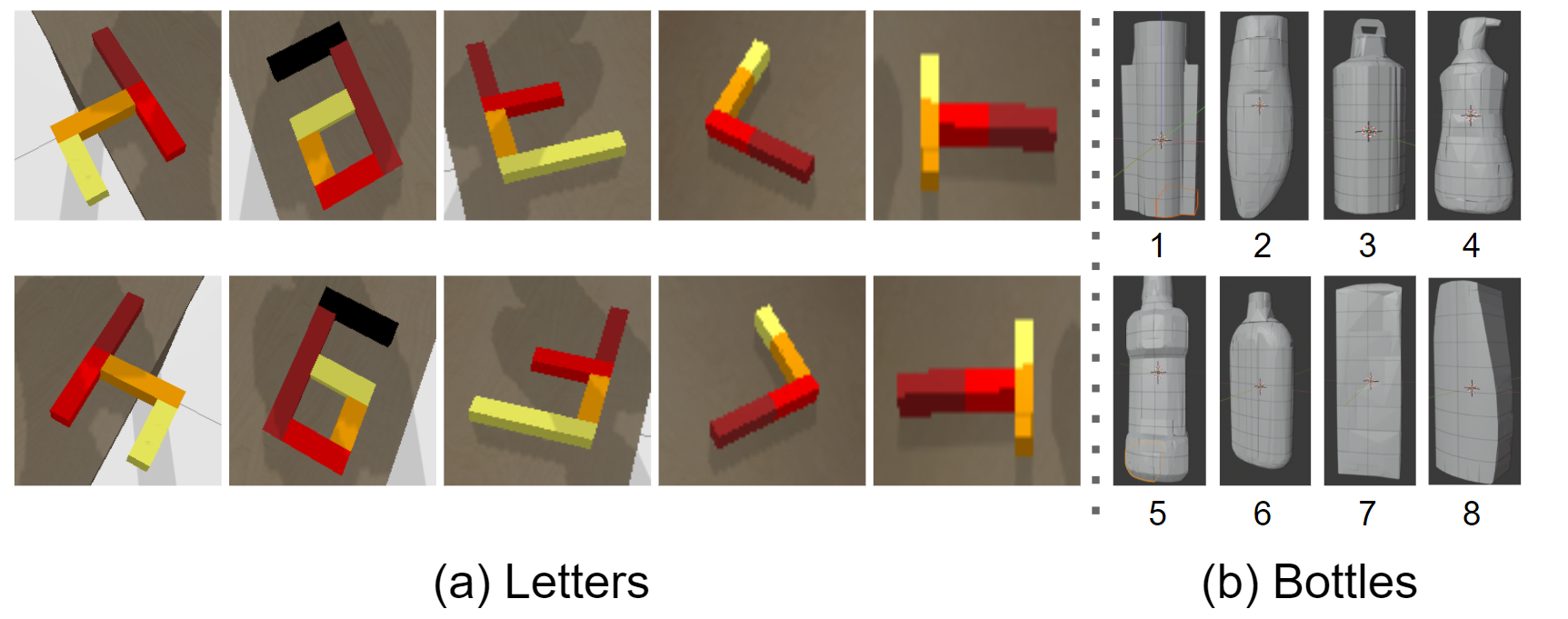}
	\caption{We generate two types of datasets. Each shape is considered as one category incorporating varying combinations of mass distribution, friction coefficient, and size. Different colors represent distinct physical properties for visualization.}
	\label{fig:synthetic_datasets}
\end{figure}

\textbf{\textit{Letters} Dataset.} 
We adhere to the Unified Robot Description Format (URDF) for the synthesis of a diverse object dataset tailored for simulation. This synthesis is structured hierarchically, wherein multiple cubes combine to form a bar, and several bars merge to compose objects of distinct shapes, and each shape is considered as a category. Alterations in object shape are accomplished by varying the number of cubes within each bar, with the option to adjust the size of individual cubes (as shown in Fig.~\ref{fig:synthetic_datasets}a). The \textit{Letters} dataset encompasses 10 distinct shape categories, each containing $200 \sim 250$ object instances generated randomly, with varying physical properties by independently manipulating the mass, friction coefficient and size of each cube. For training purposes, we select 8 categories, reserving 5\% of the objects for intra-category (IC) evaluation. The remaining 2 categories are used for cross-category (CC) evaluation.

\textbf{\textit{Bottles} Dataset.} 
This dataset encompasses 8 distinct bottle objects as shown in Fig.~\ref{fig:synthetic_datasets}b, each of which is disassembled into multiple smaller components. These object configurations are exported in XML format, with specific physical properties assigned to each component. Notably, we adopt a systematic approach, consistently assigning higher mass density and friction coefficient to one side of the object, employing scales ranging between $0.8$ and $1.2$ relative to the original object's size. Consequently, each object is treated as a category, encompassing a total of 84 variants that encompass different sizes and diverse physical property configurations.

\subsection{Simulator}
In our study, we utilize both Pybullet and Mujoco as integral components of our data collection and grasping behavior generation processes. This is motivated by our goal to comprehensively assess the distinctions between these simulators, thereby mitigating any potential biases stemming from simulator-specific effects. Our work involves the rigorous evaluation of these simulators to provide a more robust and balanced understanding of their respective performance characteristics and their implications for our research.
\begin{figure}[t]
	\centering
		\centering
		\includegraphics[width=0.48\textwidth]{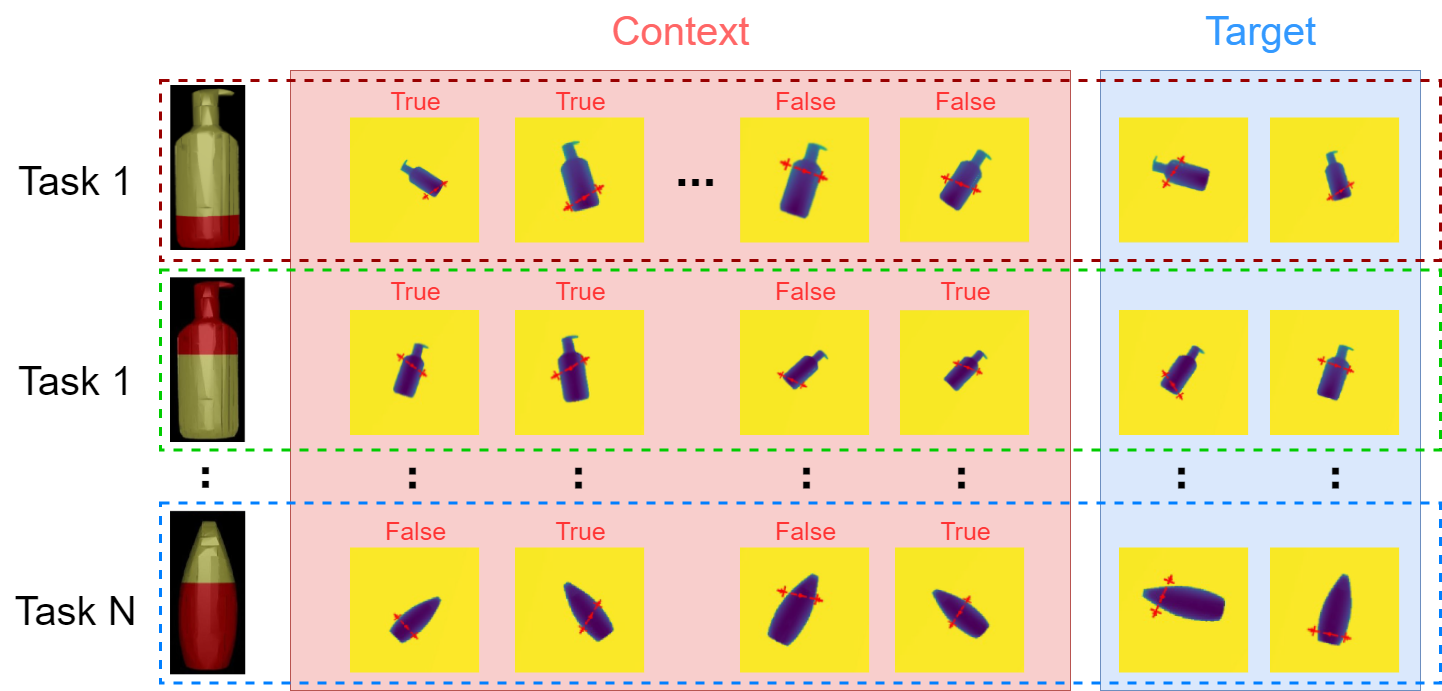}
	\caption{The dataset is split into context and target sets for each object. The context set includes the depth images $x$ w.r.t. the grasp candidates, the distance $z$ between the grasp candidates and the gripper, and the binary grasp labels indicating if the grasp succeeds. In contrast, the target set lacks labels during the inference phase. The data is split randomly between context and target sets for each training iteration.}
	\label{fig:data_split}
\end{figure}
\subsection{Meta-Learn Physical Properties as Task Embeddings}
We now formally describe ConDex in the context of meta-learning grasping. 
We assume that all objects are sampled from the same distribution $p(T)$, each object $T_i$ includes a context set of grasping observations $D_C^i=\{({x}_{C,1},z_{C,1},y_{C,1}),\dots,({x}_{C,K},z_{C,K},y_{C,K})\}_i$ and a target set $D_T^i=\{({x}_{T,1},z_{T,1},y_{T,1}),\dots,({x}_{T,M},z_{T,M},y_{T,M})\}_i$ where K and M are the number of samples in each set which could be varied at each iteration. Similar to DexNet-2.0~\cite{Mahler2017}, variables ${x}$, $z$ and $y$ are I) the cropped depth image w.r.t. the grasp candidate, ii) the distance between the grasp point and gripper and iii) the binary grasp label indicating if the grasp succeeds or not, respectively. The label of target set is used to calculate loss during training but not available during evaluation. An example is shown in Fig.~\ref{fig:data_split}.

The entire traning dataset is denoted as $D=\{D_C^i, D_T^i\}_{i=1}^N$ where N is the number of objects sampled for training. During inference, the model is tested on a new object $T^*\sim p(T)$ given a small context set, from which it has to infer a new function $f^*: (D_C^*,({x}_T^*, z_T^*)) \rightarrow \hat{y}_T^*$. In meta-learning, there are two types of learned parameters. The meta-parameters $\theta$, which is learned during a meta-training phase using $D$, and the task-specific parameters $\phi^*$ which is updated based on samples from each individual new task $D_C^*$ conditioned on $\theta$. Predictions can be constructed as $\hat{y}_T^* = f_{\theta,\phi^*}({x}_T^*, z_T^*)$, where $f$ is the meta-model parameterized by $\theta$ and $\phi^*$.

\begin{figure}[htb!]
	\centering
	\includegraphics[width=0.45\textwidth]{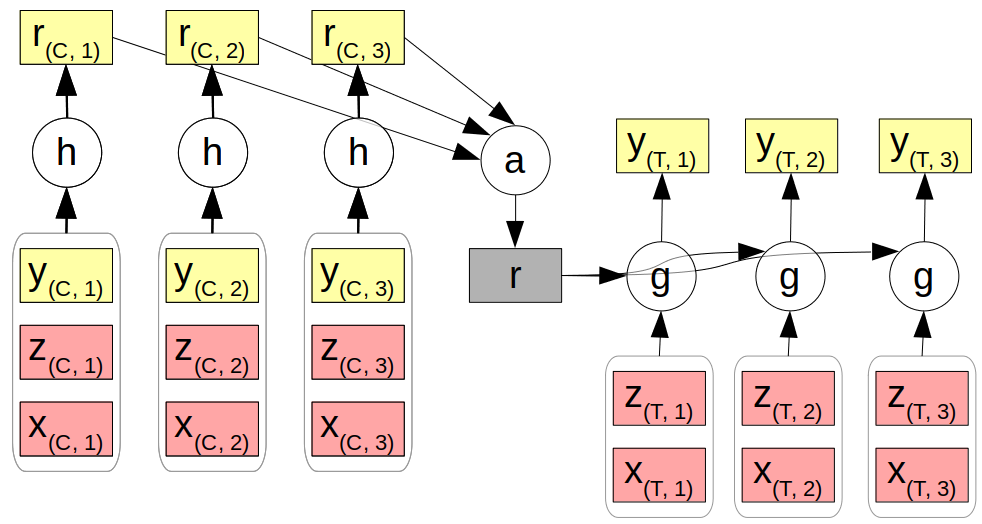}
	\caption{The structure of ConDex.}
	\label{fig:encoder-decodder}%
\end{figure}%
\begin{figure*}
    \centering
    \includegraphics[width=0.95\textwidth]{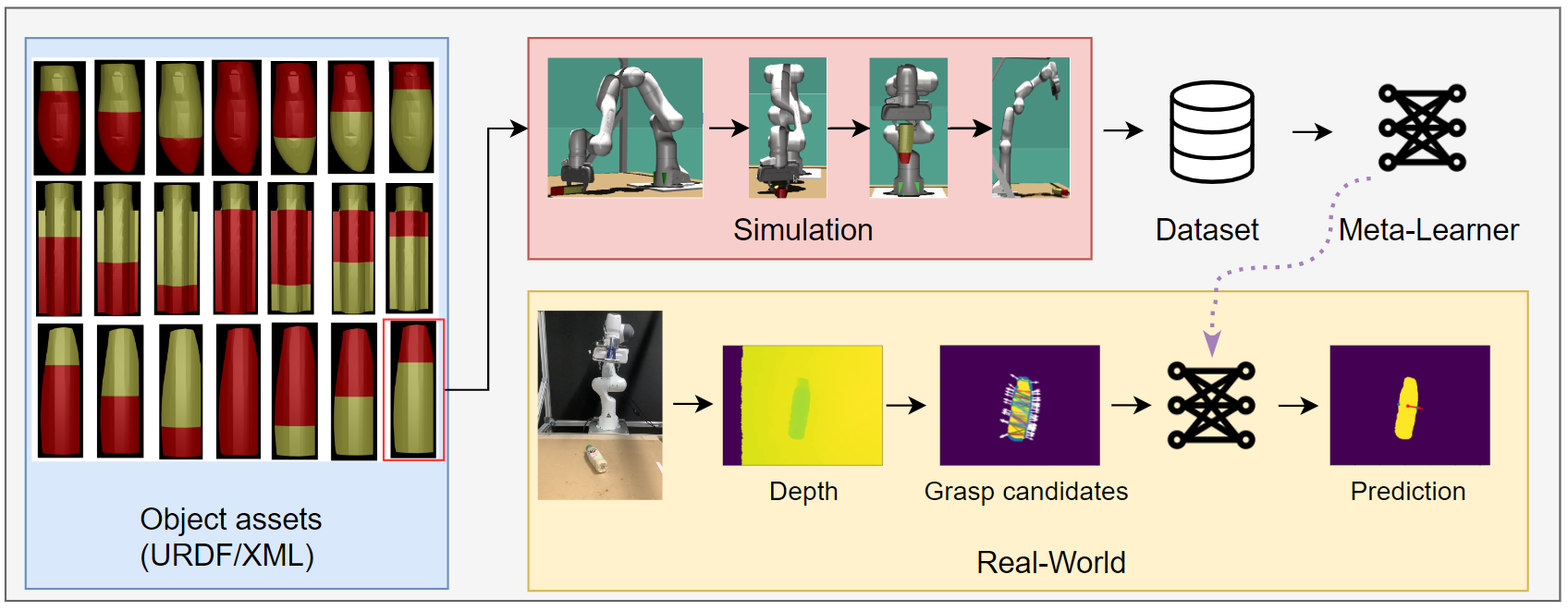}
    \caption{pipeline}
    \label{fig:pipeline}
\end{figure*}
ConDex considers $\theta$ as the neural weights consisting of an encoder $h_\theta$ and a decoder $g_\theta$. The structure of the encoder and decoder is shown in Fig.~\ref{fig:encoder-decodder}. 
$\phi^*$ is considered as the encoded task representation $\textbf{r}$ of object properties predicted from the context set of the novel task.
The encoder $h$ takes each observation consisting of the cropped depth image ${x}_{(C, i)}$, the grasping distance $z_{(C, i)}$, and the grasp label $y_{(C, i)}$ as context input and extracts the feature embedding $\textbf{r}_{(C, i)}$. 
A permutation invariant aggregator $\bigoplus$ merges all feature embeddings as a task representation $\textbf{r}$ to represent the object properties: $\textbf{r}=\bigoplus_{i=1}^K h_\theta({x}_{C,i}^*,z_{C,i}^*,y_{C,i}^*)$. Subsequently, a decoder $g_\theta$ takes $\textbf{r}$ as an additional input and outputs the score of the grasp candidate $\hat{y_T^*}=g_\theta(({x}_T^*,z_T^*),\textbf{r})$. Meta-parameter $\theta$ is fixed after training and only $\textbf{r}$ is updated as a task representation which is able to adapt in an online fashion. The network architecture is subject to the design of DexNet-2.0 except for having additional input which needs to be concatenated, i.e., the context and the task representation. We show the details of the network architecture in the supplementary video. 

The goal of ConDex is to predict the confidence score $\hat{y}_T\in [0,1]$ over the possible grasping candidates for the target set by minimizing the cross-entropy loss function:
\begin{equation}
	\theta^{*} = \mathop{\arg \min}_{\theta}\frac{1}{TM}\sum_{t=1}^{T}\sum_{i=1}^{M}L(y_{t, i}, \hat{y}_{t,i})
	\label{eq: loss-function}	
\end{equation}
where $\theta$ is the neural weights, $y$ is the ground-truth binary grasp label, $L$ is the cross-entropy loss, $T$ is the batch size of sampled tasks and $M$ is the size of the target set.

%% file: docs/experiments.tex
\section{EXPERIMENTS}
In this section, we undertake a comprehensive examination of the dataset we have gathered, offering insights into its characteristics. Subsequently, we introduce and provide a detailed comparative analysis of our proposed method alongside the established baselines. Finally, we present the outcomes of our experimentation, shedding light on the effectiveness and performance of our approach in relation to these baselines.

\subsection{Evaluation Metrics}

\textbf{Grasp Error Rate.} The inference can be considered as a binary classification. The grasping quality $Q \in [0, 1]$ indicates the confidence in each grasping candidate's success. 
The error rate is formulated as:
\begin{equation}
	Error\ Rate = \frac{FP + FN}{P + N},
	\label{eq: error-rate}	
\end{equation}
where $FP, FN$ are the number of false positives and false negatives. $P, N$ denote the successful and failed grasps in total.

\textbf{Grasping Accuracy.} We evaluate the whole grasping pipeline in both simulation and real-world scenarios and define a successful grasp when an object is lifted and dropped in the target position as shown in Fig.~\ref{fig:pipeline}. The grasping performance is indicated by the grasp accuracy:
\begin{equation}
	Grasp\ Accuracy = \frac{Number\ of\ Successful\ Grasp Trials}{Total\ Number\ of\ Grasp Trials}
	\label{eq: grasp-success-rate}	
\end{equation}
\subsection{Statistics of the Collected Data}
\begin{figure}[htb!]
	\centering
		\includegraphics[width=0.85\columnwidth]{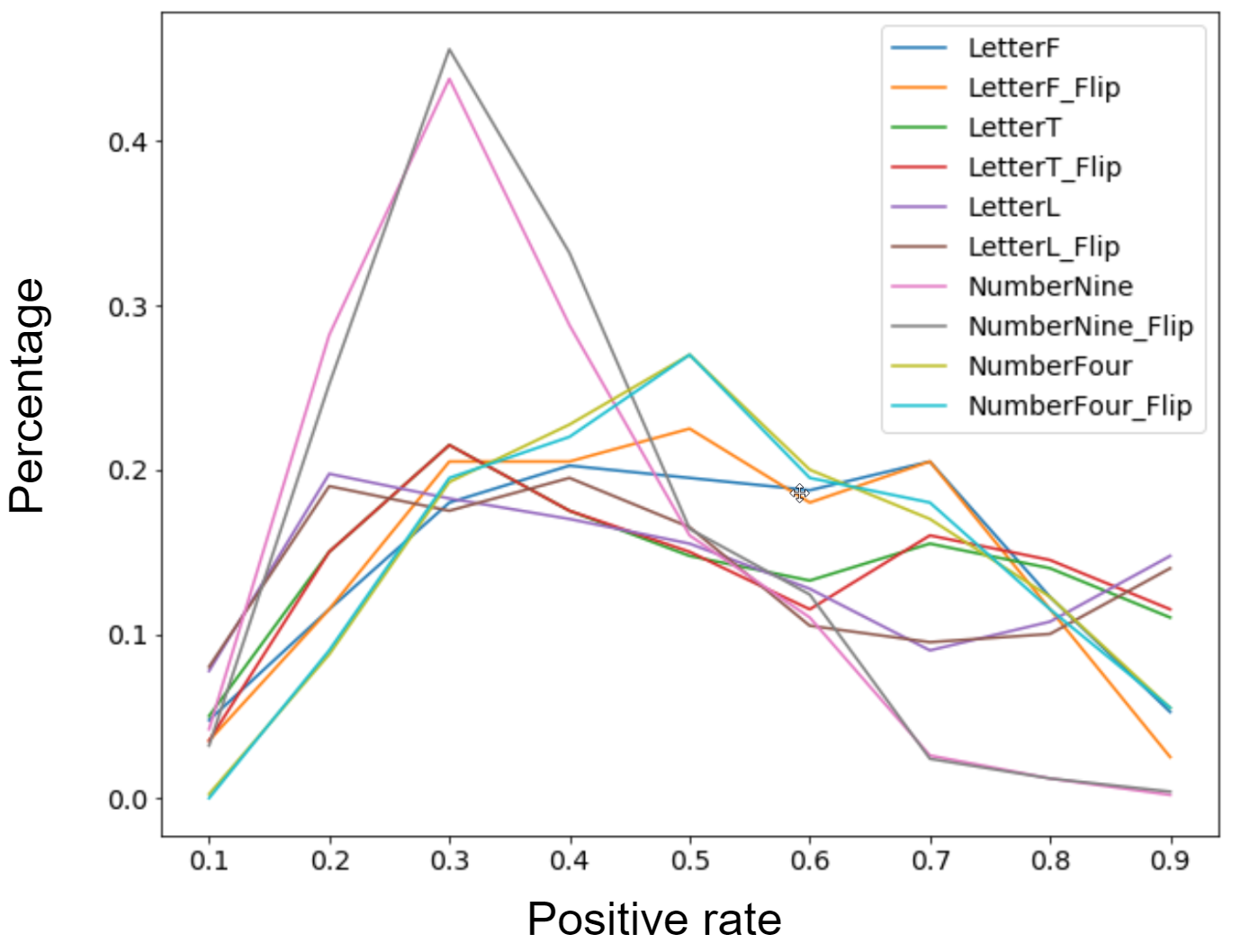}
    \caption{\textbf{Statistics of \textit{Letters} dataset.} Each curve represents one category comprising numerous instances. The collected data exhibits a normal distribution centered around a positive rate of approximately 0.5, indicating its high data quality}
	\label{fig:data_statistics}
\end{figure}
As mentioned earlier, \textit{Letters} dataset comprises 10 distinct shape categories, each containing either $200 \sim 250$ object instances. Every object instance undergoes 30 random grasps in Pybullet executed at positions within the robot arm's feasible range.
We collect in total 63000 observed data on 2.100 objects. As shown in Fig.~\ref{fig:data_statistics}, the horizontal axis represents the positive rate resulting from 30 random grasps for each object instance while the vertical axis illustrates the percentage of object instances belonging to a specific shape category. The distribution of object instances across different positive rates demonstrates a notable diversity within our collected image dataset. This diversity is shape complexity-dependent, with more intricate shape categories, such as Category Nine, exhibiting significantly lower overall positive rates in contrast to other categories. Nevertheless, our collected image data remains balanced, with a positive rate approaching 50\%.

\subsection{Baselines}
During evaluation, we employ the following baselines and two different variants of ConDex with different use cases:
\begin{itemize}
    \item \textbf{DexNet (Pretrained)} indicates a DexNet-2.0 model pretrained on an object dataset with homogeneous physical properties following ~\cite{Mahler2017}.
    \item \textbf{DexNet} indicates a DexNet-2.0 model trained only on our datasets.
    \item \textbf{IGML} is a meta-learning grasping method inspired by ~\cite{IGML} which employ MAML~\cite{maml} to learn the distinguishable grasps. We adapt this model by using depth images as input and training it on our dataset for fair comparison.
    \item \textbf{ConDex (accumulated)} indicates a ConDex model trained on our inhomogeneous dataset, where the collection of the next context is iteratively given and depends on previously acquired knowledge, which can be formed as :
    \begin{equation}
    \begin{aligned}
    	(\textbf{x}_{C,t+1},z_{C,t+1}) \sim P(&\textbf{x}_{C,t+1},z_{C,t+1}|(\textbf{x}_{C,1},z_{C,1},y_{C,1}),\dots, \\
    	&(\textbf{x}_{C,t},z_{C,t},y_{C,t}))
    \end{aligned}
    \label{eq: accumulative-collection}	
    \end{equation}
    \item \textbf{ConDex} denotes a ConDex model trained on our inhomogeneous object dataset, where context points are randomly collected. 
\end{itemize}

\subsection{Experimental Results}
\begin{figure}[htb!]
	\centering
		\includegraphics[width=0.8\columnwidth]{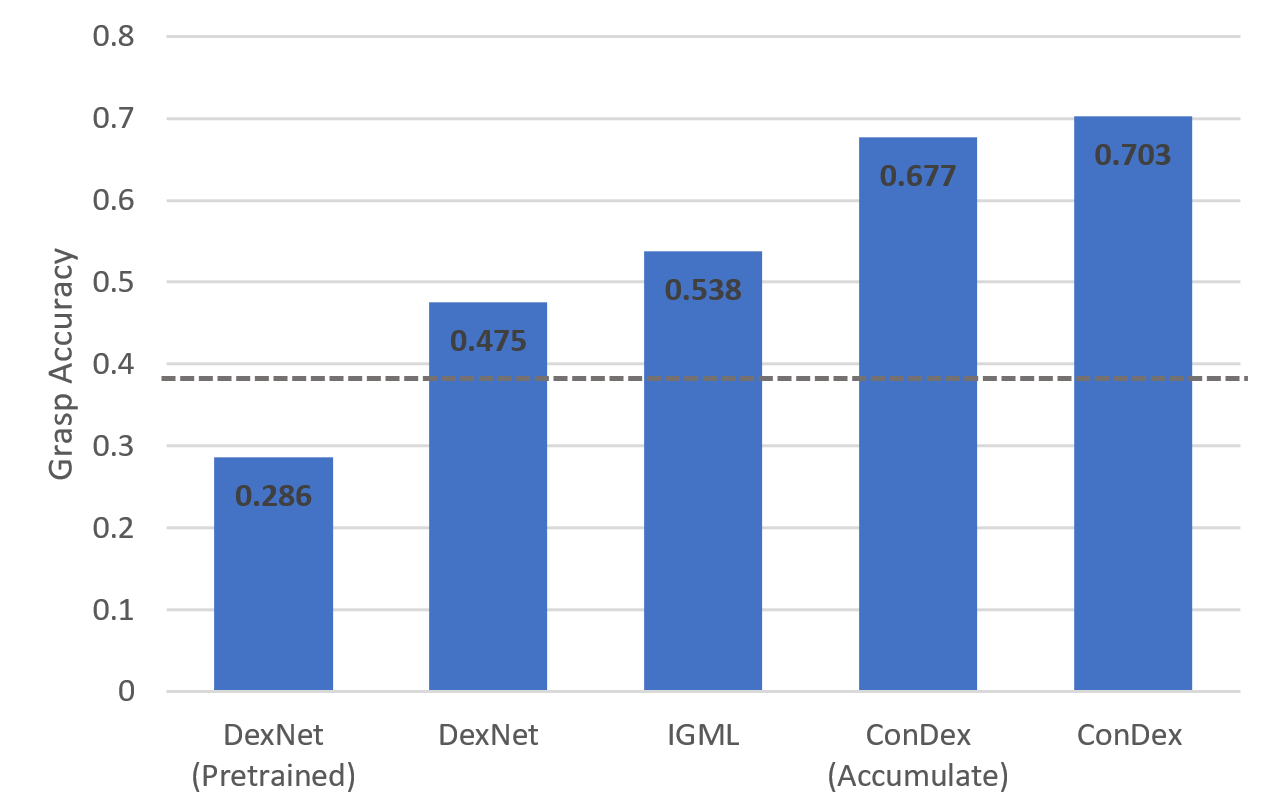}
    \caption{Results are evaluated over 50 objects from intra- and cross-categories, each object is grasped 30 times. The dashed line denotes the performance with random grasping.}
	\label{fig:exp_letters_comparison}
\end{figure}

\begin{figure}[htb!]
	\centering
		\includegraphics[width=0.85\columnwidth]{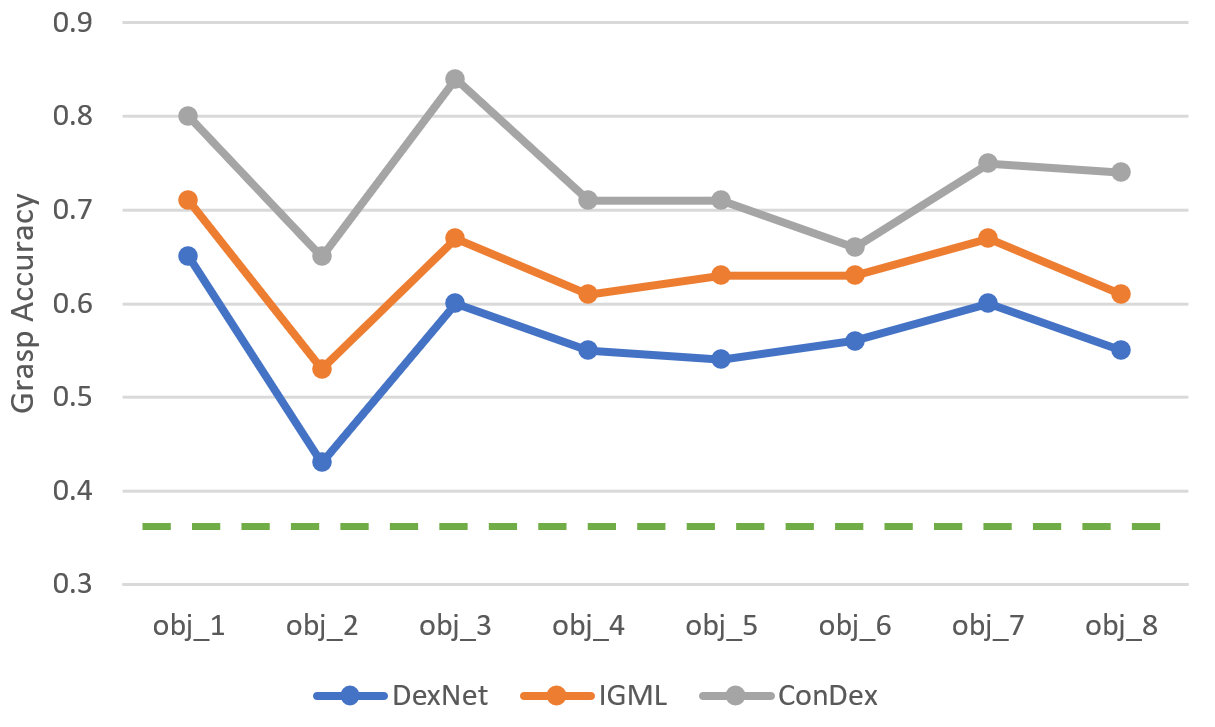}
    \caption{Evaluation on \textit{Bottles} dataset from intra- and cross-categories (i.e., object \textit{2} and \textit{6}). The dashed line denotes the performance with random grasping.}
	\label{fig:exp_syn_bottles}
\end{figure}

Fig.~\ref{fig:exp_letters_comparison} illustrates that both ConDex variants exhibit superior performance compared to the baseline approaches. Notably, DexNet (pretrained) performs even worse than random grasping, highlighting the limitation of training on large-scale homogeneous object datasets for our specific task. Conversely, these findings underscore the importance of incorporating physical properties during the training process. Fig.~\ref{fig:exp_syn_bottles} shows the evaluation on \textit{Bottles} dataset. Despite a decrease in performance observed when dealing with cross-category objects (object \textit{2} and \textit{6}) decreases, ConDex consistently outperforms the baseline methods by a substantial margin. 
\begin{figure}[htb!]
	\centering
		\includegraphics[width=0.85\columnwidth]{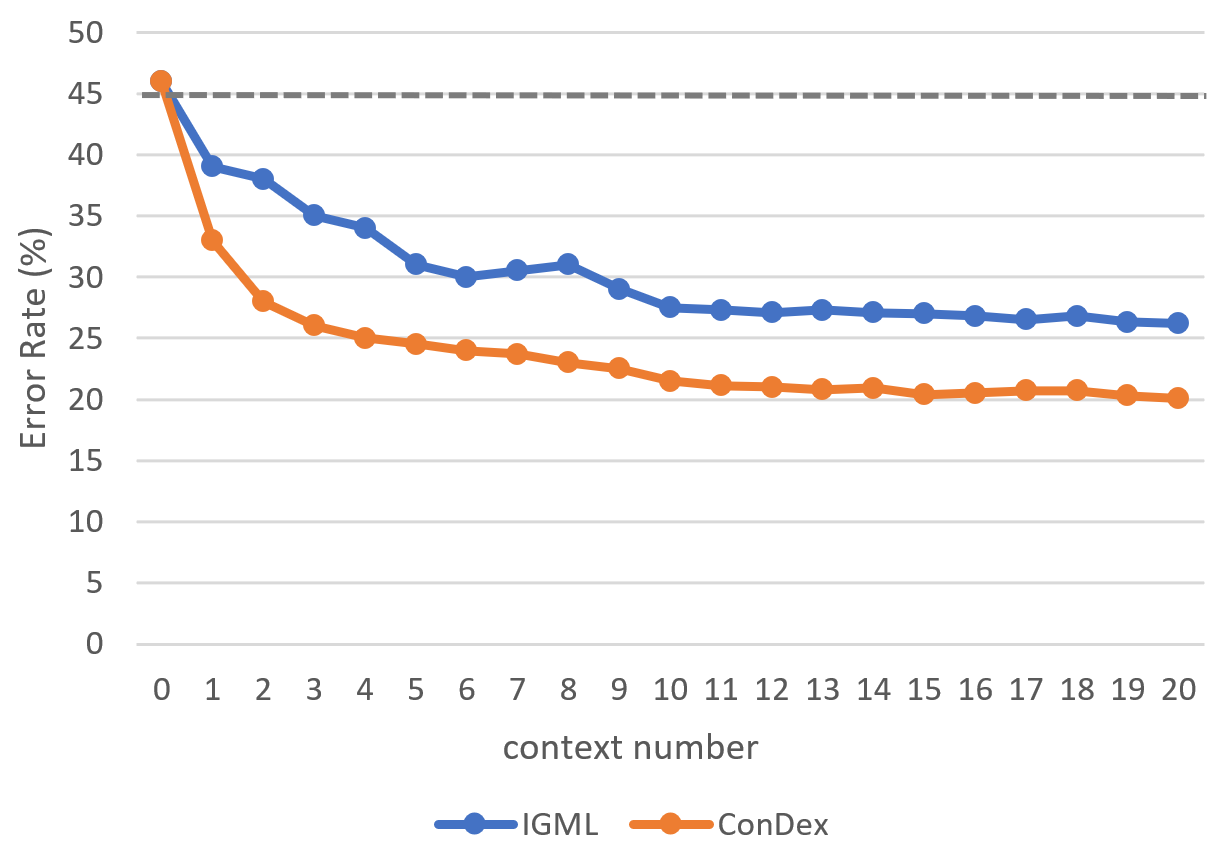}
    \caption{\textbf{Error rate vs. context number on cross-category.} 450 object instances are evaluated from two previously unseen categories from \textit{Letters} dataset. Results are presented with a maximum of 20 context points during evaluation, while a maximum of 15 context points is provided during training. The dashed line denotes the performance of DexNet.}
	\label{fig:error_rate_vs_context_letters}
\end{figure}
\begin{figure*}
	\centering
		\includegraphics[width=1\textwidth]{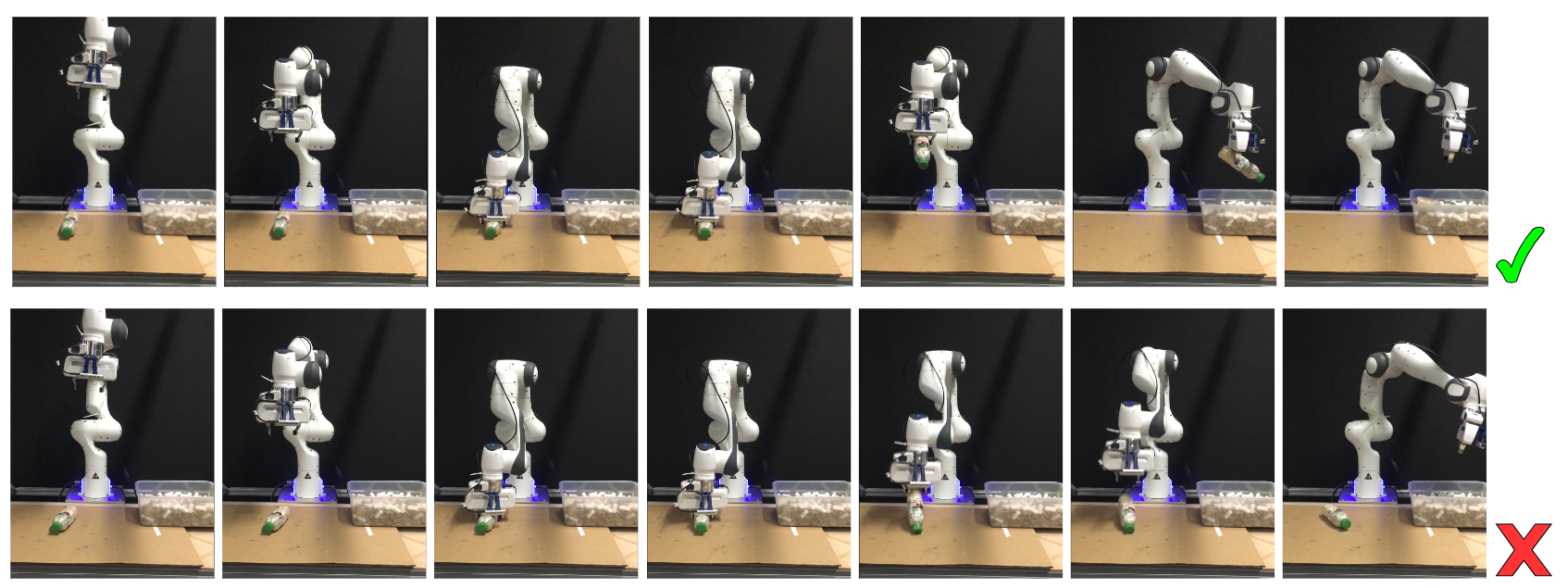}
    \caption{Conducting experiments on a real robot involves evaluating the success of manipulation based on the criteria that a successful grasp results in the bottle being successfully placed inside the designated box. The bottle is filled with different quantities of material to acquire diverse mass distributions.}
	\label{fig:real_examples}
\end{figure*}
Fig.~\ref{fig:error_rate_vs_context_letters} demonstrates the error rate with respect to the size of the context set for prediction on unseen categories. The dashed line corresponds to the performance of DexNet, which predicts independently of context information. Notably, ConDex consistently outperforms both DexNet and IGML over all different context numbers.
Furthermore, the error rate of ConDex decreases as more context points are incorporated, which indicates that the aggregation of additional contexts yields valuable information from diverse context pairs, enabling the model to adapt effectively to previously unseen tasks. Moreover, the model's performance can be further enhanced when the size of the context set exceeds the maximum number utilized during training, which was 15 in our case. Additionally, we observe a notable reduction in the error rate when the first 5 context points are provided, indicating a crucial factor in the initial stages to alleviate task ambiguity.
\begin{figure}[htb!]
	\centering
		\includegraphics[width=1\columnwidth]{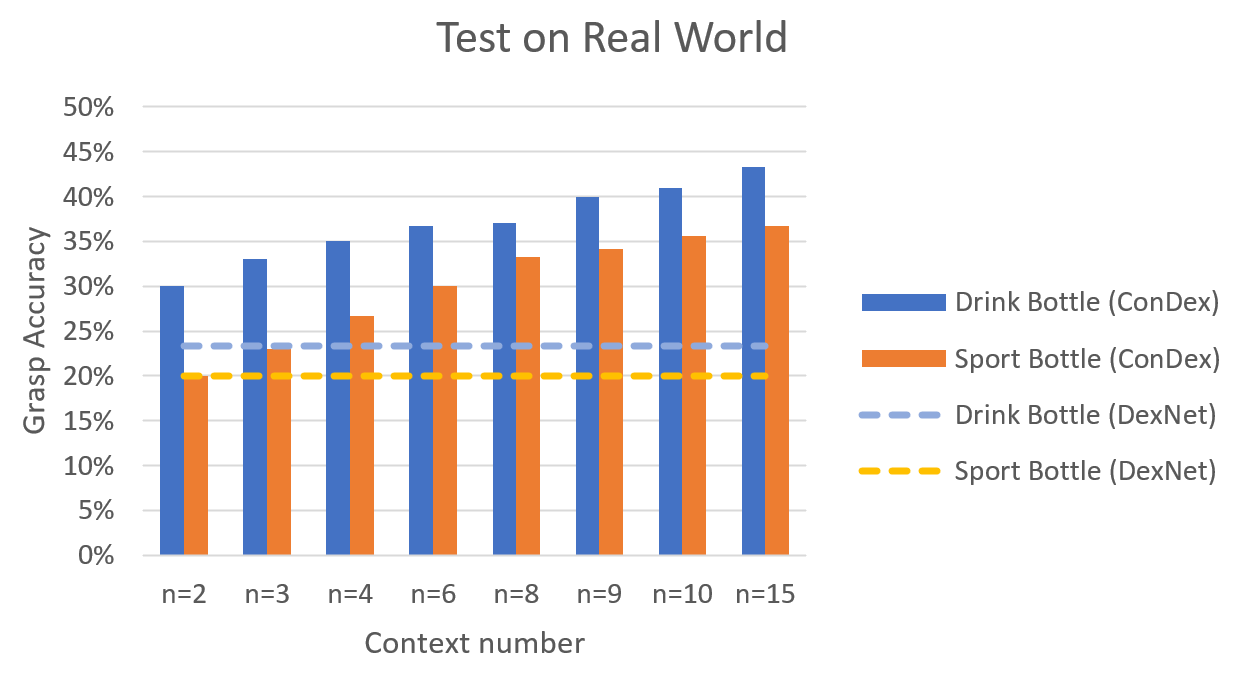}
    \caption{\textbf{Sim-to-real evaluation.} The results stem from experiments conducted on a real robot, with the models being entirely trained using the \textit{Bottles} dataset obtained through rollouts within the Mujoco simulation.}
	\label{fig:acc_vs_context_number}
\end{figure}

\textbf{Experiments on sim-to-real.} 
To assess the sim-to-real performance, we employ ConDex which is fully trained using the \textit{Bottles} dataset from Mujoco, and subsequently test on our real robot. The evaluation is conducted on two distinct types of bottles, each containing varying quantities of material. Given the differences in mass distribution among these objects, achieving a precise grasp at the correct position is crucial for successfully picking up and placing the bottle into the designated box. The experimental setup is depicted in Fig.~\ref{fig:real_examples}. The manipulation is deemed as successful if the object can be effectively placed into the box. Fig.~\ref{fig:acc_vs_context_number} presents the results with varying numbers of provided context points, along with 30 grasp executions for each trial. The enhanced performance of ConDex with the increasing number of context points signifies the successful transfer of knowledge, demonstrating the model's ability to efficiently extract valuable information from contexts. It is worth highlighting that ConDex outperforms DexNet, albeit with a slight decrease in performance compared to its performance in the Mujoco simulation.

\subsection{Limitations}
\textbf{Static physical properties:} Our approach is currently limited in its ability to handle dynamic physical properties, such as liquids with viscosity that change over time. This constraint arises from the limitations of the simulation techniques we employ. \textbf{Exclusion of transparent objects:} Transparent objects are not within the scope of our current study. Our work focuses on capturing heterogeneous properties across objects and utilizes depth as the primary input. Nonetheless, there is potential for future research to explore the inclusion of dynamic physical properties and expand the scope to encompass transparent objects as simulation techniques continue to evolve.


%% file: docs/conclusion.tex
\section{CONCLUSIONS}

In this paper, we investigate grasping challenging objects with heterogeneous physical properties using meta-learning. Due to the lack of available datasets and the relatively unexplored nature of this field, we generate two datasets encompassing diverse mass distributions and friction coefficients, collecting data from both Pybullet and Mujoco simulation environments. These datasets are crucial in evaluating the effectiveness of our proposed model ConDex against baselines. Our study underscores the significance of leveraging contextual information, facilitating fast adaptation to complex objects and enabling seamless sim-to-real transfer. We hope that our research highlights the potential of this emerging direction and raises further attention in the field.